\documentclass[letterpaper]{article} 
\usepackage{aaai24}  
\usepackage{times}  
\usepackage{helvet}  
\usepackage{courier}  
\usepackage[hyphens]{url}  
\usepackage{graphicx} 
\usepackage{natbib}  
\usepackage{caption} 
\usepackage{algorithm}
\usepackage{algorithmic}

\usepackage{arydshln}
\usepackage{multirow}
\usepackage{multicol}
\usepackage{bm}
\usepackage{url}
\usepackage{underscore}
\usepackage{amsfonts}
\usepackage{makecell}
\usepackage{enumitem}
\usepackage{booktabs}
\usepackage{subfigure}

\usepackage{newfloat}
\usepackage{listings}
\DeclareCaptionStyle{ruled}{labelfont=normalfont,labelsep=colon,strut=off} 
\floatstyle{ruled}
\newfloat{listing}{tb}{lst}{}
\floatname{listing}{Listing}
\title{Towards Efficient Diffusion-Based Image Editing with Instant Attention Masks}
\author{
    Siyu Zou\textsuperscript{\rm 1}\thanks{This work was done during her internship at Fuxi AI Lab.}\equalcontrib, 
    Jiji Tang\textsuperscript{\rm 2}\equalcontrib, 
    Yiyi Zhou\textsuperscript{\rm 1}, 
    Jing He\textsuperscript{\rm 1}, 
    Chaoyi Zhao\textsuperscript{\rm 2},\\ 
    Rongsheng Zhang\textsuperscript{\rm 2}, 
    Zhipeng Hu\textsuperscript{\rm 2}, 
    Xiaoshuai Sun\textsuperscript{\rm 1}\thanks{Corresponding author}\\
}
\affiliations{
    \textsuperscript{\rm 1} Key Laboratory of Multimedia Trusted Perception and Efficient Computing, \\ Ministry of Education of China, Xiamen University, 361005, P.R. China\\
    \textsuperscript{\rm 2} Fuxi AI Lab, NetEase Inc., Hangzhou, China\\

    zousiyu@stu.xmu.edu.cn, tangjiji01@corp.netease.com, zhouyiyi@xmu.edu.cn, blinghe@stu.xmu.edu.cn \\ \{zhaochaoyi, zhangrongsheng, zphu\}@corp.netease.com, xssun@xmu.edu.cn
}

\usepackage{cite}

\begin{document}

\maketitle

\begin{abstract}
Diffusion-based Image Editing (DIE) is an emerging research hot-spot, which often applies a semantic mask to control the target area for diffusion-based editing. However, most existing solutions obtain these masks via manual operations or off-line processing, greatly reducing their efficiency. In this paper, we propose a novel and efficient image editing method for Text-to-Image (T2I) diffusion models, termed \emph{Instant Diffusion Editing} (InstDiffEdit). In particular, InstDiffEdit aims to employ the cross-modal attention ability of existing diffusion models to achieve instant mask guidance during the diffusion steps. To reduce the noise of attention maps and realize the full automatics, we equip InstDiffEdit with a training-free refinement scheme to adaptively aggregate the attention distributions for the automatic yet accurate mask generation.
Meanwhile, to supplement the existing evaluations of DIE, we propose a new benchmark called \emph{Editing-Mask} to examine the mask accuracy and local editing ability of existing methods. To validate InstDiffEdit, we also conduct extensive experiments on \emph{ImageNet} and \emph{Imagen}, and compare it with a bunch of the SOTA methods.
The experimental results show that InstDiffEdit not only outperforms the SOTA methods in both image quality and editing results, but also has a much faster inference speed, \emph{i.e.}, +5 to +6 times. Our code available at \url{https://github.com/xiaotianqing/InstDiffEdit}
\end{abstract}

\section{Introduction}

For a year or two, diffusion models have gradually become the mainstream paradigm in conditional image generation ~\cite{saharia2022photorealistic, ramesh2022hierarchical, rombach2022high, balaji2022ediffi, nichol2021glide}. Compared with \emph{Generative Adversarial Networks} (GAN) ~\cite{karras2019style, karras2020analyzing, xia2021tedigan}, diffusion models yield a completely different generation pipeline, which can obtain more diverse and interpretable generations. The great success of diffusion models also sparks researchers to apply them to the task of semantic image editing~\cite{meng2021SDEdit, kawar2022imagic}.

\begin{figure} [tbp]
\centering
\includegraphics[width=1\linewidth]{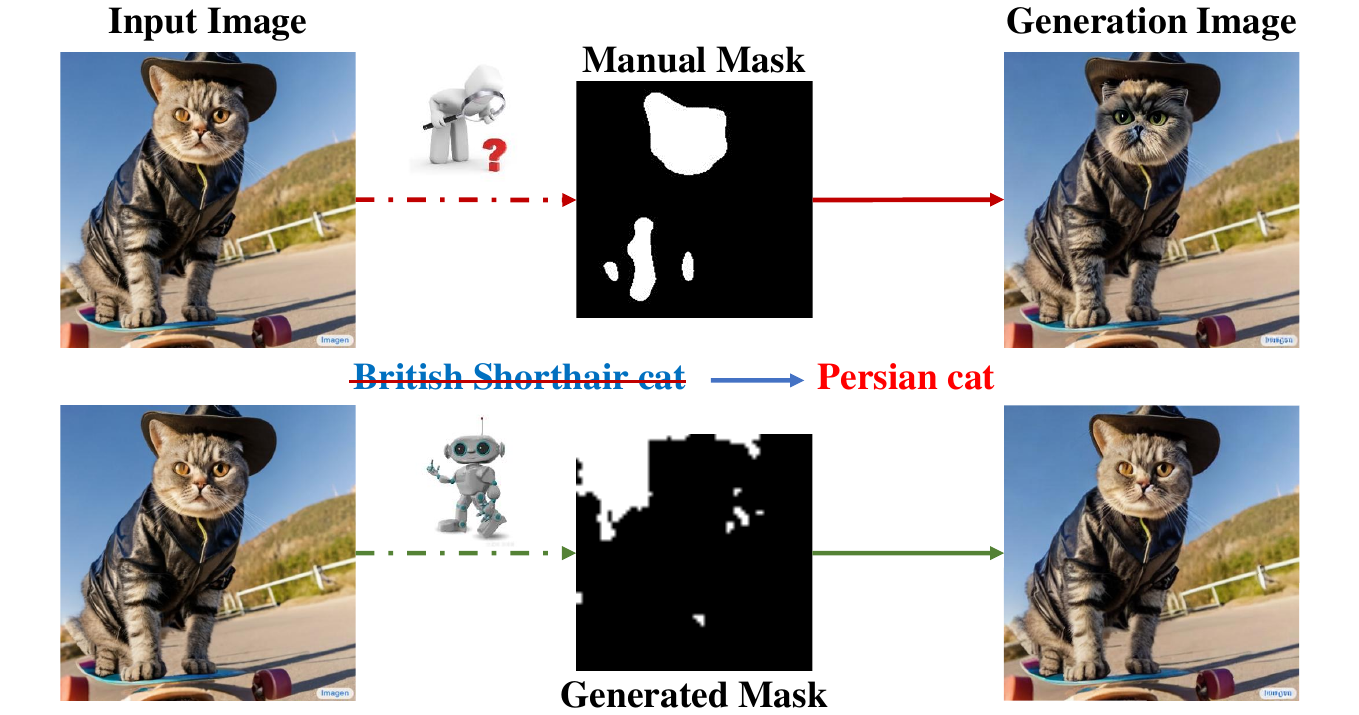} 
\caption{Illustration of existing diffusion-based image editing methods, where a manually or off-line generated mask is often used to control the editing area.}
\label{fig1}
\end{figure}

Semantic image editing ~\cite{zhan2021multimodal} aims to modify the target instance of the given image according to the input text description, while the rest image information needs to be preserved as much as possible. Although existing diffusion models ~\cite{saharia2022photorealistic, ramesh2022hierarchical, rombach2022high} excel in generation quality and diversity on text-to-image generation, it still lacks precise controls. Therefore, recent diffusion-based editing methods introduce additional information to better control the image manipulation, such as reference image ~\cite{meng2021SDEdit} or semantic mask ~\cite{avrahami2022blended1}. 

Among these solutions, padding a semantic mask is the most effective way for accurate image editing, which can precisely restrict the target image area and achieve editing via text-to-image diffusions~\cite{avrahami2022blended1}, as shown in Fig. \ref{fig1}. However, the mask generation often requires manual intervention~\cite{avrahami2022blended1, couairon2022DiffEdit}, greatly limiting the efficiency of these methods for the practical use.

Recent advance has aspired to automate the editing process via reducing the manual efforts or including the mask generation in diffusion models. For instance, PtP ~\cite{hertz2022prompt} proposes a semi-automated method, which can directly obtain mask by manually setting some parameters. More recently, DiffEdit~\cite{couairon2022DiffEdit} proposes a fully automatic method, which can embed the mask generation into the diffusion framework, but its mask generation and image editing are still time consuming. Overall, existing solutions still exhibit obvious shortcomings in terms of either manual intervention or computation efficiency. 

In this paper, we propose a novel yet efficient image editing method for diffusion models, termed \emph{Instant Diffusion Editing} (InstDiffEdit). The feasibility of InstDiffEdit is attributed to the superior cross-modal alignment of existing diffusion models. In the advanced diffusion models like Stable Diffusion~\cite{rombach2022high}, an effective multi-modal space has been well established by learning numerous image-text pairs, and these models also involve excellent cross-attention mapping. In this case, we can leverage the hidden attention maps in diffusion steps to facilitate instant mask generation. However, these hidden attention maps are intractable to directly use, and they are often full of noise. For instance, the semantic attentions of start token are much more noisier than that of ``\emph{cat}'' in Fig. \ref{vision}. Thus, we also equip InstDiffEdit with a learning-free mask refinement scheme, which can adaptively aggregate the attention distributions according to the editing instruction. Notably, the proposed InstDiffEdit is a plug-and-play component for most diffusion models, which is also training-free.

To validate InstDiffEdit, we apply it to Stable Diffusion v1.4~\cite{rombach2022high}, and conduct extensive experiments on two benchmark datasets, namely ImageNet~\cite{deng2009ImageNet} and Imagen~\cite{saharia2022photorealistic}. Meanwhile, to better measure the local editing ability and mask accuracy of existing methods, we also propose a composite benchmark called \emph{Editing-Mask}, as a supplementary evaluation to DIE. The experimental results on ImageNet and Imagen show that compared with existing methods, InstDiffEdit can achieve the best trade-off between computation efficiency and generation quality for semantic image editing. For instance, compared with the recently proposed DiffEdit, our method can obtain competitive editing results while improving the inference speed by 5 to 6 times. The results on Editing-Mask confirm the superiority of our method in background preservation. Furthermore, we also provide sufficient visualizations to examine the ability of InstDiffEdit.

Conclusively, the contribution of this paper is three-fold:
\begin{itemize}
 \item We propose a novel and efficient image editing method for diffusion-based models, termed \emph{InstDiffEdit}, which obtains instant mask guidance via exploiting the cross-modal attention in diffusion models. 
 \item As a plug-and-play component, InstDiffEdit can be applied to most diffusion models for semantic image editing without further training or human intervention, and its performance is also SOTA.
 \item We propose a new image editing benchmark, termed Editing-Mask, containing 200 images with human-labeled masks, which can be used for the evaluation of mask accuracy and local editing ability.
\end{itemize}

\section{Related Work}
\subsection{Text-to-Image Diffusion}
\begin{figure} [t]
\centering
\includegraphics[width=1\linewidth]{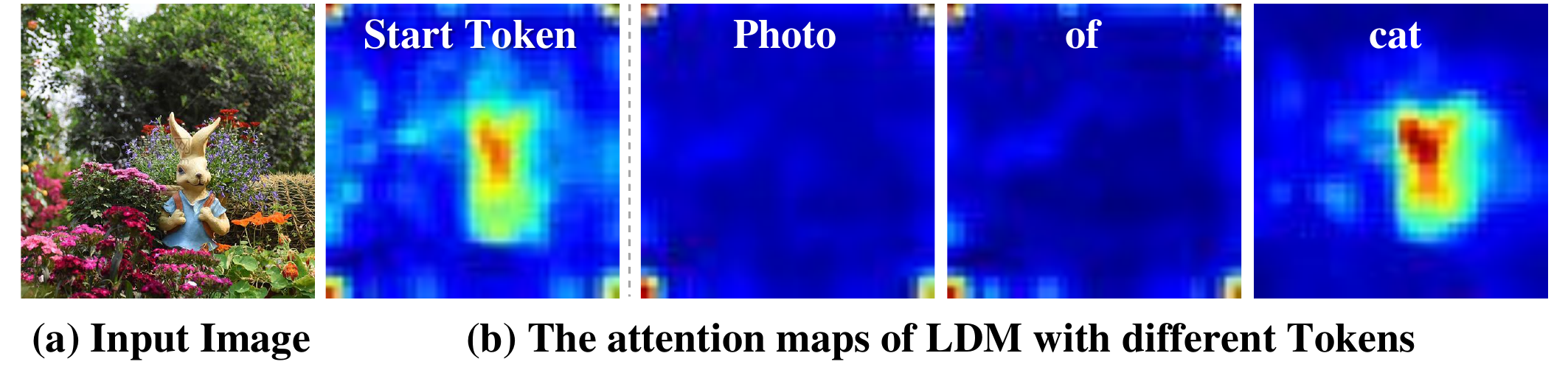} 
\caption{ The visualization of the attention maps in Stable Diffusion. The target word of ``\emph{cat}'' has the best attention map, but it needs to be manually identified during applications. The start token is relevant but still very noisy.} 
\label{vision}
\end{figure}

In the past few years, a lot of diffusion-based methods~\cite{rombach2022high, ramesh2022hierarchical, saharia2022photorealistic} has been proposed, which also demonstrate superior performance in terms of image quality and diversity compared to GAN.~\cite{karras2020analyzing, xia2021tedigan}. Some recent works~\cite{avrahami2022blended1} also explore the combination of diffusion models with \emph{Contrastive Language-Image Pre-Training} (CLIP)~\cite{radford2021learning}. For example, Stable Diffusion ~\cite{rombach2022high} leverages CLIP's text encoder to guide the image generation process. By incorporating cross-attention between text and noisy images, the model generates images that are semantically aligned with the textual description. 

\subsection{Semantic Image Editing}

A plethora of GAN-based semantic image editing approaches~\cite{gan,xu2018attngan, xia2021tedigan} have been proposed with remarkable outcomes. The emergence of large-scale GAN networks, such as the StyleGAN family~\cite{karras2019style, karras2020analyzing, karras2021alias}, significantly enhances the editing capabilities. Meanwhile, Transformer~\cite{vaswani2017attention} has demonstrated remarkable performance in text-driven image editing tasks. ManiTrans~\cite{wang2022manitrans} use Transformers to predict the content of covered regions, which enables semantic editing only performing on a certain image region. 

Recently, with the developments of diffusion models, practitioners also explore their application in semantic image editing. 
SDEdit~\cite{meng2021SDEdit} accomplishes this by retaining a portion of the reference image information during the diffusion process. CycleDiffusion~\cite{wu2022unifying} proposes an inversion model to get a better latent from the input image, thus improving the edit quality. PtP~\cite{hertz2022prompt} and PnP~\cite{tumanyan2022plug} operate editing via modifying attention maps in diffusion models. More recently, to prevent unbounded edits from global image editing, some methods resort to local editing techniques. For example, Blended Diffusion~\cite{avrahami2022blended1} and RePaint~\cite{lugmayr2022repaint} implement local editing on real images with manual mask. 
However, the acquisition of manual masks is time-consuming and labor-intensive, and hinders the developments of automated semantic editing.

\begin{figure*} [htp]
\centering
\includegraphics[width=1\linewidth]{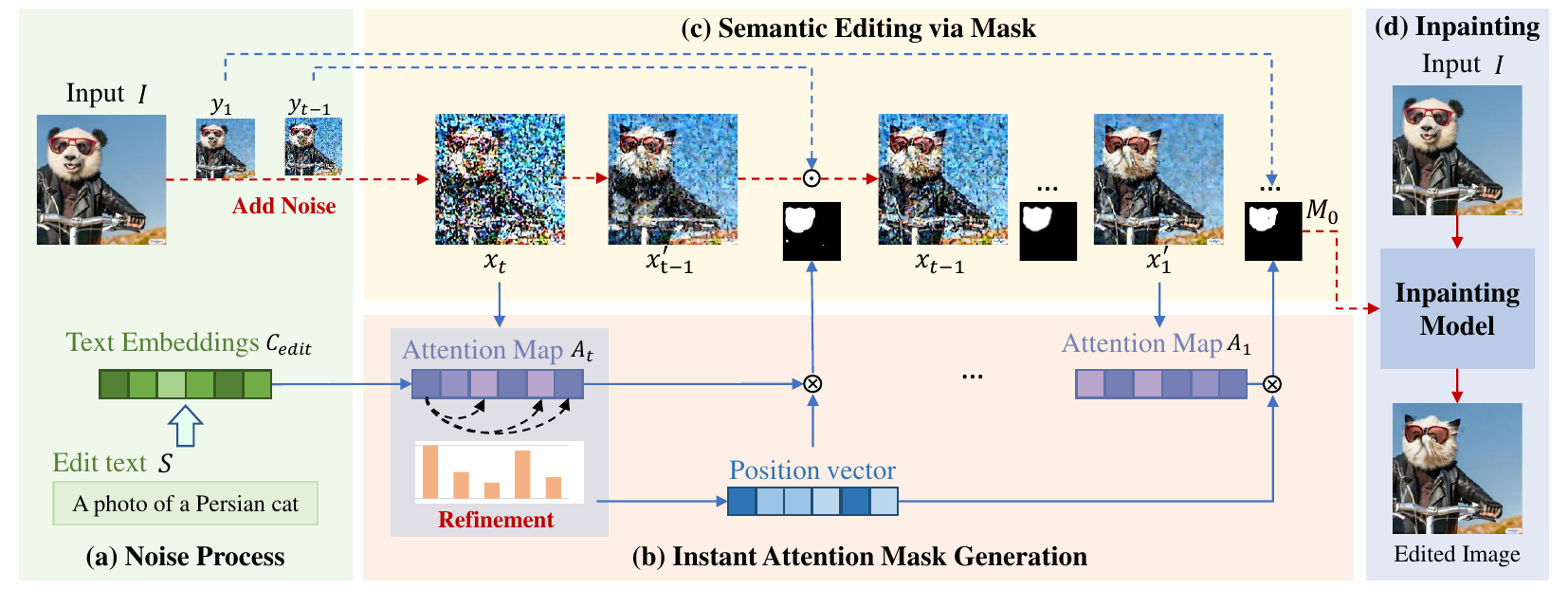} 

\caption{ The framework of the \emph{Instant Diffusion Editing} (InstDiffEdit). InstDiffEdit involves instant mask generation at each denoising step based on the attention maps. This mask can provide instant guidance for the image denoising. The left part (a) illustrates the noise process, and (b) depicts the generation of semantic mask at each step, based on which the diffusion-based image editing is performed (c). Lastly, the inpainting model is further applied to accomplish the generation (d). } 

\label{overviews}
\end{figure*} 

Therefore, some methods have begun to explore automated mask generation. 
DiffEdit ~\cite{couairon2022DiffEdit} is better suited to the requirements of automated editing as it obtains the mask by contrasting variations in model predictions with different text prompts. However, because of the stochastic randomness of the diffusion model, DiffEdit requires multiple iterations to stabilize the ultimate output, which leads to inefficiencies in terms of time.

\section{Preliminary}
\subsection{Latent Diffusion Models}
Traditional diffusion models~\cite{ho2020denoising} typically operate the diffusion process on high-resolution image space, which significantly limits training and generation speed. In order to achieve more efficient training and generation, Latent Diffusion Models (LDMs)~\cite{rombach2022high} perform the diffusion process on the latent space rather than the resolution space, thereby improving the efficiency of training and inference. 

First of all, LDMs leverages an automatic encoder framework $E_I$, such as VAE~\cite{kingma2013auto}, to map the image features $I$ to low-dimensional latent spaces $x_0$ and generate noisy image features $x_t$ through the diffusion forward process:
\begin{equation} \label{eq:addnoise}
x_t=\sqrt{\overline{\alpha}_t}x_0+\sqrt{1-\overline{\alpha}_t}\epsilon_t, x_{0} = E_I(I), 
\end{equation} 
where $t$ denotes the time-step, which is determined by noise strength $r$. The noise term $\epsilon_{t}$ is sampled from a \emph{standard normal distribution}. $\alpha_{t}$ is a decreasing schedule of diffusion coefficients that controls the strength of noise ateach step.

Subsequently, the text sequence $S$ is mapped to a feature space using a text encoder $E_T$ such as CLIP~\cite{radford2021learning}, recorded as $C_{edit}= E_T(S)$. The diffusion process period is operated on latent space, denoted as:
\begin{equation} 
x_{t-1} = \frac{1}{\sqrt{\alpha_t}}(x_t - \frac{1-\alpha_t}{\sqrt{1-\overline{\alpha}_t}}\epsilon_\theta(x_t, c, t)) + \sigma_tz.
\end{equation} 

Finally, a decoder $D_I$, which corresponds to the encoder $E_I$, is employed to reconstruct the image from the latent dimension with $I_{rec} = D_I(x_{0})$.

\subsection{Cross-Attention in LDMs}
In LDMs, text-to-image generation is accomplished by modifying the latent representations using cross-attention alignments. Specifically, for each text $S$ which consists of N tokens, the pre-trained text encoder $CLIP_T$ is utilized to transform it into the text feature $c=\{c_1, c_2, \dots, c_N\}$. Similarly, input image is transformed into image latent $x_0$ and the noisy image latent $x_t$ is obtained according to Eq. \ref{eq:addnoise}.

Subsequently, the text features and image latent are projected by three trainable linear layers, denoted as $f_Q$, $f_V$, and $f_K$. 
Next, the spatial attention maps $A$ is generated for each text token by:
\begin{equation} \label{eq.crossattention}
\small
A=Softmax(\frac{QK^T}{\sqrt{d_k}}), Q=f_Q(z_t), K=f_K(t), V=f_V(t)
\end{equation} 
where $d_k$ denotes the feature dimension of K. And the attention maps $A$ is then combined with the value matrix $V$ to obtain the final output of the cross-attention layer with $V\cdot A$.

Generally, the attention maps in Stable Diffusion can indicate the correspondence between text words and image regions. 
However, due to the noise contained in image latent , it is challenging to directly obtain the desired target instance from the attention maps, and these hidden attention maps are still of noisy, as shown in Fig. \ref{vision}.

\section{Methodology}

\subsection{Overview}
In this paper, we propose a novel and efficient image editing method based on text-to-image diffusion models, termed \emph{Instant Diffusion Editing} (InstDiffEdit), of which structure is illustrated in Fig. \ref{overviews}. 

Concretely, similar to exisitng methods~\cite{avrahami2022blended1}, we aim to achieve the target image editing by padding a semantic mask to input image, based on which the diffusion steps are conducted to achieve target edition. This process can be defined by:
\begin{equation} 
x_t= M \cdot {x}'_t + (1-M) \cdot y_t, \label{eq.blendmask}
\end{equation} 
where, ${x}'_t$ and $y_t$ denote the predicted noisy latent and the latent representation of the noisy image at step $t$, and $M$ is the mask. Then, we can get the noisy latent $x_t$ for editing.

This mask-based editing is supported by recent advances in diffusion models~\cite{avrahami2022blended1}, which can restrict editing areas using mask and replace the non-masked area of the predicted image with noise image at the current timestep. This allows mask-based methods to preserve the background in the non-masked area while editing. However, the generation of this semantic mask often requires manual efforts ~\cite{hertz2022prompt, patashnik2023localizing} or off-line processing~\cite{avrahami2022blended1, lugmayr2022repaint}. In this case, InstDiffEdit resorts to the attention maps in LDMs for instant mask genernation during diffusions. As shown in Fig. \ref{vision}, the attention maps in LDMs capture the semantic correspondence between the image and text well. 

However, it also encounters some problems. To specify the attention map of the editing target, \emph{e.g., ``cat"} in Fig. \ref{vision}, the method still requires manual efforts, since we do not know the length and content of user's instruction during application. And directly using the map of \emph{``start token"} as a trade-off is still too noisy for efficetive edition. 

In this case, we equip InstDiffEdit with an automatic refinement scheme for mask generation. As shown in Fig. \ref{overviews}, given an input image latent feature $x_t$, and a text feature $C_{edit}$, we can get the hidden attention maps $A$ in denoise process from Eq. \ref{eq.crossattention}. Then, we propose a parameter-free attention mask generation module $G( \cdot )$ to obtain the semantic mask $M_t=G(x_t, C_{edit})$. Later, with this instant mask, we can directly perform target image editing during the diffusion steps, which can be re-written by:
\begin{equation}
    x_{t-1}=M_t \cdot \epsilon_\theta(x_t, t, C_{edit})+(1-M_t) \cdot y_t.
\end{equation}
where, $M_t$ is the mask computed by the attention mask module in timestep $t$ and $\epsilon_\theta$ denotes the diffusion model. 

Lastly, in order to achieve better generation results, we adopt a strategy of using the mask generated in the last denoising step as the final mask, and generating the final editing results through the inpainting way in LDMs.

In the next subsection, we will give the detail definition of the proposed attention mask generation module.

\begin{figure} [t]
\centering
\includegraphics[width=1\linewidth]{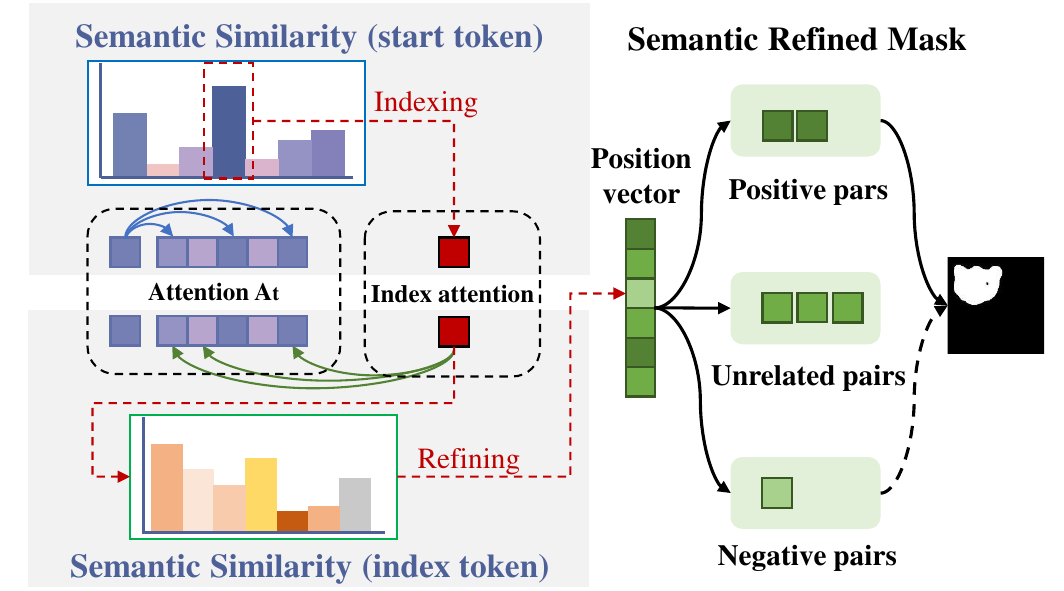} 

\label{fig:mask}
\caption{ The proposed instant mask generation. An indexing process is first performed based on the semantic similarities between the start token and the other ones (upper left). Refinement is then operated between the index and the remaining ones (lower left). Finally, the mask is obtained via the adaptive aggregation of all attention maps.} 
\end{figure}
\subsection{Instant Attention Mask Generation}
In InstDiffEdit, we use the attention maps generated in the denoising process as the information source for mask generation. However, the input text often consists of multiple tokens, and the attention information of each token has its own focus and varies vastly with the change of sentence length and word composition. Therefore, it is difficult for the model to automatically locate attention results of the target words.

In practice, we use the attention maps of the start token as the base information for further attention mask refinements. To explain, in a well pre-trained T2I diffusion model, the start token often expresses the semantics of the whole sentence. As shown in Fig. \ref{vision}, the focus region of attention corresponding to the start token overlaps highly with the edit region of the semantic description. 
However, the start token contains the whole sentence as well as part of the original image information, so its attention distribution is still messy. 

In this case, we adopt the idea of key information extraction to eliminate the noisy information and obtain the most relevant content with semantic information. 
Assuming a noise strength of $r$, the denoise process starts at time-step $\tau$ ($\tau=r*T, T=1000$), and the corresponding attention maps $A_{\tau}$ can be obtained using Eq. \ref{eq.crossattention}. Specifically, we leverage the attention map of the start token $A_{start}^{\tau} \in R^{16 \times 16}$ as the reference information, and subsequently retrieve the attention $A_{index}^{\tau}\in R^{16 \times 16}$ by computing all similarities with the reference map. This enables us to identify the location of the object that requires modification:
\begin{equation} 
A_{index}^{\tau}=argmax\sum_{i\in[1, N]}cosine(A_i^{\tau}, A_{start}^{\tau}), 
\end{equation} 
where $cosine(\cdot)$ denotes semantic similarity and $N$ is the length of all tokens in sentence.

To obtain more accurate mask information, we further aggregate the concept-related information and eliminate irrelevant information. Specifically, we compute the similarities between the obtained $A_{index}^{\tau}$ and the attention maps of the text tokens to obtain a similarity vector $S\in R^{1\times N}$:
\begin{equation} 
S_{i}=cosine_{i\in[1, N]}(A_i^{\tau}, A_{index}^{\tau}).\label{eq.refine1}
\end{equation} 

In principle, the similarity of the attention maps at each token is closely related to the semantic similarity of the sentence. As the attention maps are associated with the core semantic, the similarities will be larger, and \emph{vice versa}.

Afterwards, we can get a position vector to weight the attention information via filtering the similarity vector with two thresholds:
\begin{equation}
\label{eq.refine2}
\small
    P_{i\in [1, N]}=\left\{\begin{array}{l}
    1 \quad \quad S_i > \gamma_1, \\
    -1 \quad S_i < \gamma_2, \\
    0 \quad \quad others.\\ 
    \end{array}\right.
\end{equation}

Computing semantic similarities at each step of the denoising process can be time-consuming due to the large dimensionality of the attention maps. To mitigate this issue, we propose to compute the position vector $P$ only in the first step $\tau$ of the denoising process.
 
Finally, we obtain the refined attention map $A_{t}^{ref}$ with attention maps $A_t$ and $P$ at timestep $t \in \{{\tau},\dots,0\}$ ($A_t^{ref}=P \cdot A_t$), which is then processed using Gaussian filtering and \emph{binarized} with a threshold $\varphi$ to obtain the final mask $M_t$: 
\begin{equation}
    M_t(x,y)=\left\{\begin{array}{l}
    1 \quad  {A_t^{ref}}(x,y)>\varphi, \\
    0 \quad others.\\ 
    \end{array}\right.
\end{equation}
here, $(x, y)$ refers to a point in the latent space of the image.
Notably, the above instant attention mask generation module is training free, and thus it can be directly plugged into most existing T2I diffusion models. Meanwhile, through the refine processing, the obtained mask is much superior than the ones before refining.

\subsection{Semantic Editing via Mask}
Through the mask generation module, we obtain a mask at each step of the image denoising process. Thus, by blending the mask, guidance can be provided to denoising by Eq. \ref{eq.blendmask}. 

However, since all the information in the masked area is essentially discarded, the resulting image often has local semantic consistency but does not consider global semantics, leading to artifacts. Additionally, when the noise level is low, some editing operations cannot be achieved, such as color modification. Thus, we also equip InstDiffEdit with an inpainting based method for semantic image editting. 

The inpainting method~\cite{rombach2022high} initializes the information in the masked area with completely random noise and considers global information during generation, thus eliminating artifacts and editing failures caused by the original image information. Nevertheless, the performance of inpainting is highly dependent on the accuracy of mask. 

Therefore, we combine the advantages of the two methods by using attention maps to generate mask in the denoising process, thereby guiding image generation and obtaining more accurate mask during denoising. 

Finally, we use the inpainting method on the mask generated in the last step of denoising to generate an image that is artifact-free and more consistent with the remaining information in the original image. Notably, the combination of two mask editing methods only slightly increases the computation cost of semantic image editing.

\section{Experiments}
\subsection{Experiment Setting}
\subsubsection{\textbf{Datasets}}

We use ImageNet, Imagen and Editing-Mask to evaluate the performance of semantic editing task. 

\begin{itemize}
\item \textbf{ImageNet} Followed the evaluation of Flexit~\cite{couairon2022flexit}. A total of 1092 images in ImageNet~\cite{deng2009ImageNet} are included, covering 273 categories. For each image, the edit text is another similar category .
\item \textbf{Imagen} We construct an evaluation dataset for semantic editing by utilizing the generations from the Imagen~\cite{saharia2022photorealistic} model. Specifically, we randomly selected a short text which not in the input text as the edit text, such as replacing "British shorthair cat" with "Shiba Inu dog", resulting in a dataset of 360 paired samples.
\item \textbf{Editing-Mask} A new dataset, which comprises 200 images randomly selected from Imagen and ImageNet. Each sample includes an image, input text, edit text, and a human-labelled mask that corresponds to the semantics of the edit text. Our proposed dataset enables direct evaluation of the performance of editing tasks, particularly in regions where editing is necessary.
\end{itemize}

\begin{table*}[!htbp]
\centering
\resizebox{\linewidth}{!}{ 
    \setlength{\tabcolsep}{1.5mm}{}
    \begin{tabular}{cc|cccccccccccccc}
    \toprule
    \multirow{2}{*}{\textbf{Category}} &
    \multirow{2}{*}{\textbf{Models}} & \multirow{2}{*}{\textbf{Time}$\downarrow$} & \multicolumn{4}{c}{\textbf{Editing-Mask}} & & \multicolumn{2}{c}{\textbf{ImagetNet}} & &\multicolumn{3}{c}{\textbf{Imagen}} \\
    \cline{4-7}
    \cline{9-10} 
    \cline{11-14} 

    & & &\small{\textbf{IOU}}$\uparrow$ & {\boldsymbol{{$C_{m}$(\%)}}}$\uparrow$ &  \small{\boldsymbol{{$C_{non}$(\%)}}}$\downarrow$ & {\textbf{rate}}$\uparrow$  & &
    \small{\textbf{LPIPS}}$\downarrow$ &  \small{\textbf{CSFID}}$\downarrow$ & &\small
    {\textbf{LPIPS}}$\downarrow$ &  \small{\textbf{FID}}$\downarrow$ & \small{\textbf{CLIPScore}}$\uparrow$

    \\
    \midrule
    \multirow{2}{*}{\textbf{Latent}}  
    & SDEdit & 3.0 & -  & 11.6 & 8.4 & 1.38  & &
        31.1 & 76.5 & &
        32.1 &  75.2 &0.238
    \\
    & CycleDiffusion & 5.2 & - & 12.1 & 7.3 & 1.66  &&
        31.1 & 87.5 & &
        25.8 & 63.0 &  0.246
    \\

    \midrule
    \multirow{2}{*}{\textbf{Attention}}  
    
    & PtP  & 18.2 & -& 16.8 & 12.9 & 1.30 &&
        - & - & &
        42.8 & 85.67 & 0.240
    \\
    & PnP & 80.0 & -& 12.1 & 7.8 & 1.56 &&
        \textbf{27.3} & 76.8 & &
        22.2 & 61.6 & 0.240
    \\
    
    \midrule  
    \multirow{2}{*}{\textbf{Mask}}  
    & DiffEdit & 64.0 & 33.0  &  19.5 &  8.0 & 2.45  & &
        27.9 & 70.9 & &
        29.7 & 58.8 & 0.247 
    \\
    & InstDiffEdit  & 10.8 & \textbf{56.2} & \textbf{22.7} & \textbf{ 6.1} & \textbf{3.71} & &
        28.6 & \textbf{65.1} & &
        \textbf{17.0} & \textbf{55.3} & \textbf{0.249} 
    \\
    \bottomrule
   \end{tabular}
}
\caption{Comparison with existing methods on three datasets. The performance of Mask-based methods are much ahead of other methods. Moreover, InstDiffEdit leads to 70.3\% on IOU and 51.4\% on  changing rate \small{${C_m}/{C_{non}}$} compared with SOTA method DiffEdit. All experiment are conducted on a NVIDIA A100.}
\label{tab:main-results}
\end{table*}

\subsubsection{Metrics}

We evaluate the performance of editing methods in terms of time efficiency and generation quality. Specifically, we measure the average editing time of an image at a resolution of 512 to assess the time consumption of each method. Additionally, we used the \emph{Learned Perceptual Image Patch Similarity} (\textbf{LPIPS})~\cite{zhang2018unreasonable} metric to quantify the difference between the generated image and the original image, which reflects the degree of modification made by the editing method. Furthermore, we employed the \emph{Classwise Simplified Fréchet Inception Distance} (\textbf{CSFID})~\cite{couairon2022flexit} metric, which is a category FID metric that measures the distance between generated and original images.
We also use \textbf{CLIPScore}~\cite{hessel2021clipscore} to measure the semantic similarities between the edit texts and generated images.
It is noted that all of these metrics evaluate the generated image quality rather than the editing performance. Therefore, in our proposed human-labeled mask daraset, we use \emph{Intersection over Union} (\textbf{IOU}) to assess the quality of the generated masks, \boldsymbol{${C_m}$} and \boldsymbol{$C_{non}$} to represented the modifications of the image in the mask and non-mask areas. The metrics on Editing-Mask provides a more direct evaluation of editing performance.

\begin{figure*}[tp]
\centering
\subfigure{
\begin{minipage}[t]{0.25\linewidth}
\centering
\includegraphics[width=1\linewidth]{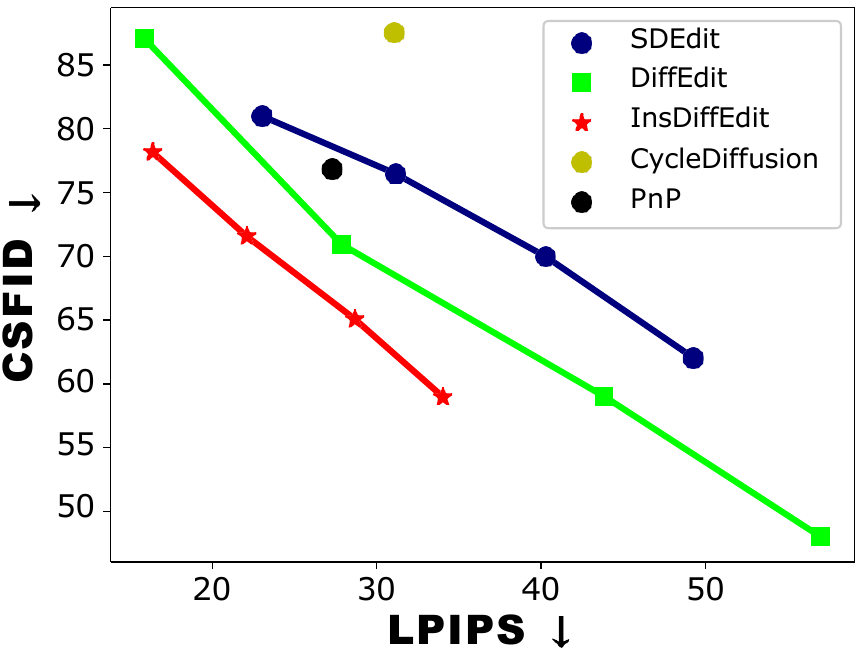} 
\end{minipage}%
}%
\hspace{10mm}
\centering
\subfigure{
\begin{minipage}[t]{0.25\linewidth}
\centering
\includegraphics[width=1\linewidth]{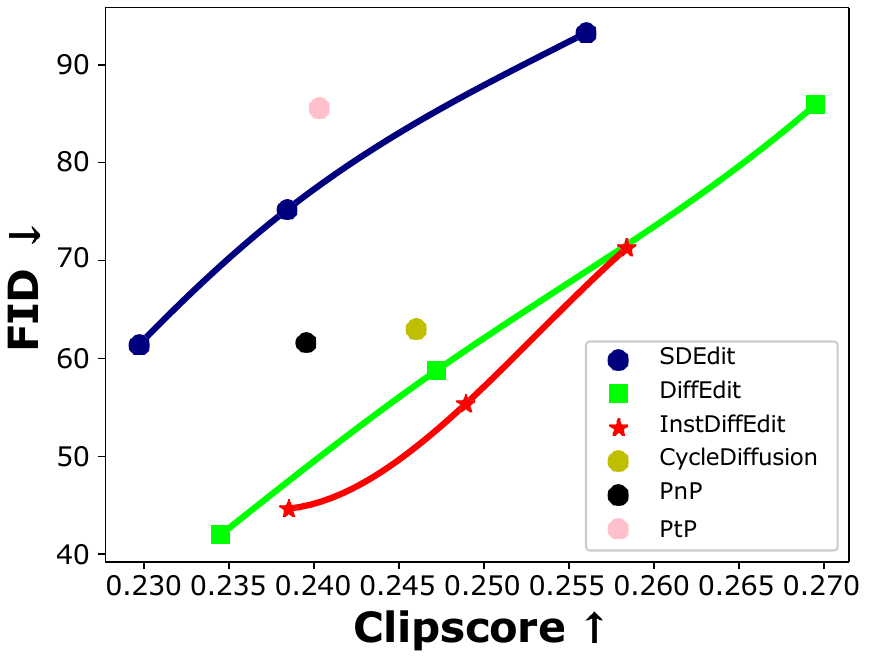}
\end{minipage}%
}%
\hspace{10mm}
\subfigure{
\begin{minipage}[t]{0.25\linewidth}
\centering
\includegraphics[width=1\linewidth]{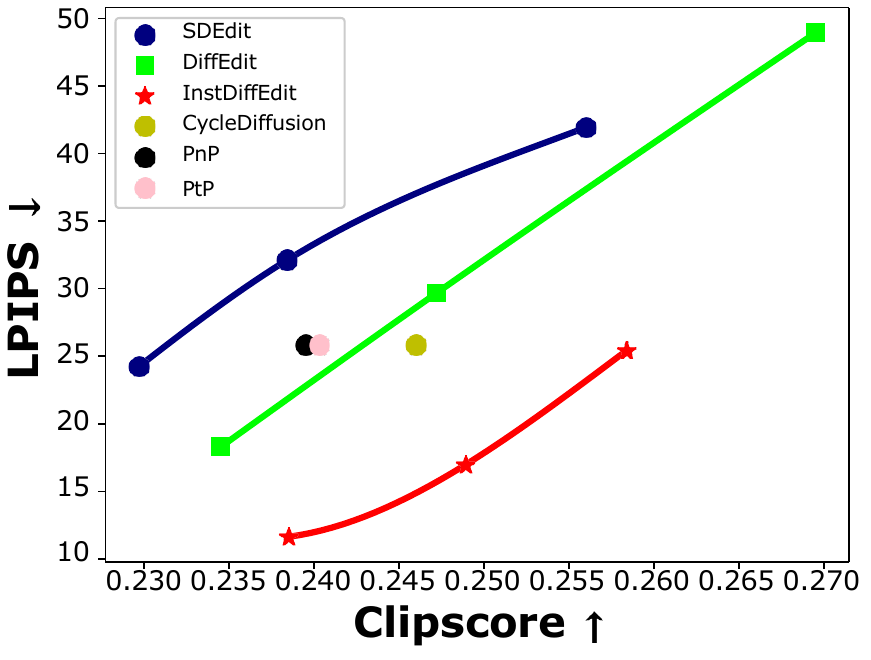}
\end{minipage}%
}%
\centering
\caption{ The trade-offs of existing methods between different metrics. We conduct experiments by using two different metrics as the independent and dependent variables respectively. The proposed InstDiffEdit has the best trade-offs. }
\label{fig.figs}
\end{figure*}

\subsubsection{Implementation}

The framework of InstDiffEdit is based on Stable Diffusion v1.4. We use 50 steps of LDMScheduler sampler with a scale 7.5, and set noise strength to $r=0.5$, threshold of binarization to $\varphi=0.2$, and the thresholds for attention refinement defined in Eq. \ref{eq.refine2} are 0.9 and 0.6 by default, respectively. We maintain $n=3$ rounds of denoising on the input image in parallel throughout the entire denoising process. Finally, we use the inpainting mode in Stable Diffusion to get the target image. 

\subsection{Experimental Results}

\subsubsection{\textbf{Quantitative Analysis}}
In this section, we present quantitative results on three datasets.

\textbf{Comparison With Existing Methods.} 
To validate the effectiveness of the proposed InstDiffEdit, we compare it with five diffusion-based methods, of which results are given in Tab. \ref{tab:main-results} and Fig. \ref{fig.figs}. The latent-based methods, \emph{i.e.}, SDEdit \cite{meng2021SDEdit} and CycleDiffusion \cite{wu2022unifying}, which rely on the association between the generated image's latent and the original image's latent. These methods offer the advantage of low time cost for editing. However, their performance is much worse than the other methods. Meanwhile, attention-based methods, \emph{i.e.}, PtP \cite{hertz2022prompt} and PnP \cite{tumanyan2022plug}, infer on the latent representation of real images, resulting in lower time efficiency and heavy reliance on the performance of inversion.
As a mask-based model, DiffEdit \cite{couairon2022DiffEdit} achieves significant improvements over all datasets, indicating the effectiveness of generated masks in diffusion-based image editing. Specifically, on our proposed Editing-Mask, DiffEdit's changing rate ${C_m}/{C_{non}}$ far exceeds that of latent-based and attention-based methods. However, DiffEdit still requires much longer inference time.
In stark contrast, our InstDiffEdit
achieves up to 5 to 6 times faster inference speeds than DiffEdit, while obtaining more accurate masks. InstDiffEdit also demonstrates improvements of IOU with ground truth masks, changing rates with 70.3\% and 51.4\%, respectively. This strongly confirms that the proposed mask generation scheme can generate more accurate masks. Results on ImageNet show that InstDiffEdit generally outperforms DiffEdit in terms of image quality, although its LPIPS score is slightly worse
. Additionally, InstDiffEdit's performance on the CSFID benchmark significantly outperforms DiffEdit by +21.1\%. Similar results are also observed on the Imagen benchmark, where InstDiffEdit excels in both image quality and image-text matching, achieving a performance increase of +44.8\% compared to DiffEdit on LPIPS.



We also depict the performance trade-offs between different metrics in Fig. \ref{fig.figs}. These results are achieved by tuning the hyper-parameters of each method based on the target metric.
From these figures, we can first conclude that the proposed InstDiffEdit can consistently achieve the best trade-offs on all metric pairs. We observe that InstDiffEdit significantly outperforms the other methods under all conditions. These results further confirm the advantages of InstDiffEdit in terms of diffusion-based image editing.

\begin{table}[t]
    \centering
    \setlength\tabcolsep{4pt}
    \renewcommand\arraystretch{0.8}
    \begin{tabular}{cc|cccc}
    \toprule
    {$\bf{r}$} & {$\bf{\varphi}$}  & {\bf IOU$\uparrow$} & {\bf {$C_m$(\%)} $\uparrow$} & {\bf {$C_{non}$(\%)}   $\downarrow$}  & {\bf {$C_{m}$/$C_{non}$}$\uparrow$} \\
    \toprule
    
    \textbf{0.5}  & None & - & 11.6 & 8.4 & 1.38 \\
    \midrule
    \multirow{3}*{\textbf{0.4}} 
      & 0.1 & 52.9 & 26.0 & 8.2 & 3.16 \\
    ~ & 0.2 & 55.7 & 21.8 & 6.0 & 3.63 \\
    ~ & 0.3 & 52.0 & 17.3 & \textbf{4.7} & 3.68 \\
    \midrule
    \multirow{3}*{\textbf{0.5}} 
      & 0.1 & 51.9 & 27.4 & 8.7 & 3.16 \\
    ~ & 0.2 & \textbf{56.2} & 22.7 & 6.1 & 3.71 \\
    ~ & 0.3 & 54.3 & 18.2 & 4.8 & \textbf{3.81} \\
    \midrule
    \multirow{3}*{\textbf{0.6}} 
      & 0.1 & 49.6 & \textbf{28.1} & 9.4 & 2.98 \\
    ~ & 0.2 & 54.6 & 24.3 & 6.7 & 3.60 \\
    ~ & 0.3 & 54.2 & 19.3 & 5.1 & 3.76 \\
    \bottomrule
    \end{tabular}
    \caption{Ablation study of noise strength $r$ and binarization threshold $\varphi$ on Editing-Mask.}
    \label{tab.ablation}
\end{table}

\textbf{Ablation Study.} 
Tab. \ref{tab.ablation} presents ablation results for different settings of the noise strength $r$ in Eq. \ref{eq:addnoise} and the binarization threshold $\varphi$. In the firs row, we assess the method's performance without a mask, and the insufficient performance indicates that mask-free methods are inferior in for image editing. Secondly, as the noise strength $r$ increases, the model obtains less information from the original image and tends to generate masks with larger areas, which results in an upward trend of $C_{m}$ and $C_{non}$ (\emph{Line 2} vs \emph{Line 5} vs \emph{Line 8}). However, the IOU with ground truth mask and change rate exhibits a trend of initially increasing and then decreasing. Additionally, as the binarization threshold $\varphi$ decreases, there is a tendency for the mask to cover a larger region, resulting in a similar phenomenon as discussed previously. Therefore, we select 
$r=0.5$ and $\varphi=0.2$, which yields the highest IOU and superior performance on the change rate.

\renewcommand{\dblfloatpagefraction}{.9}
\begin{figure*} [btp]
\centering
\includegraphics[width=0.95\linewidth]{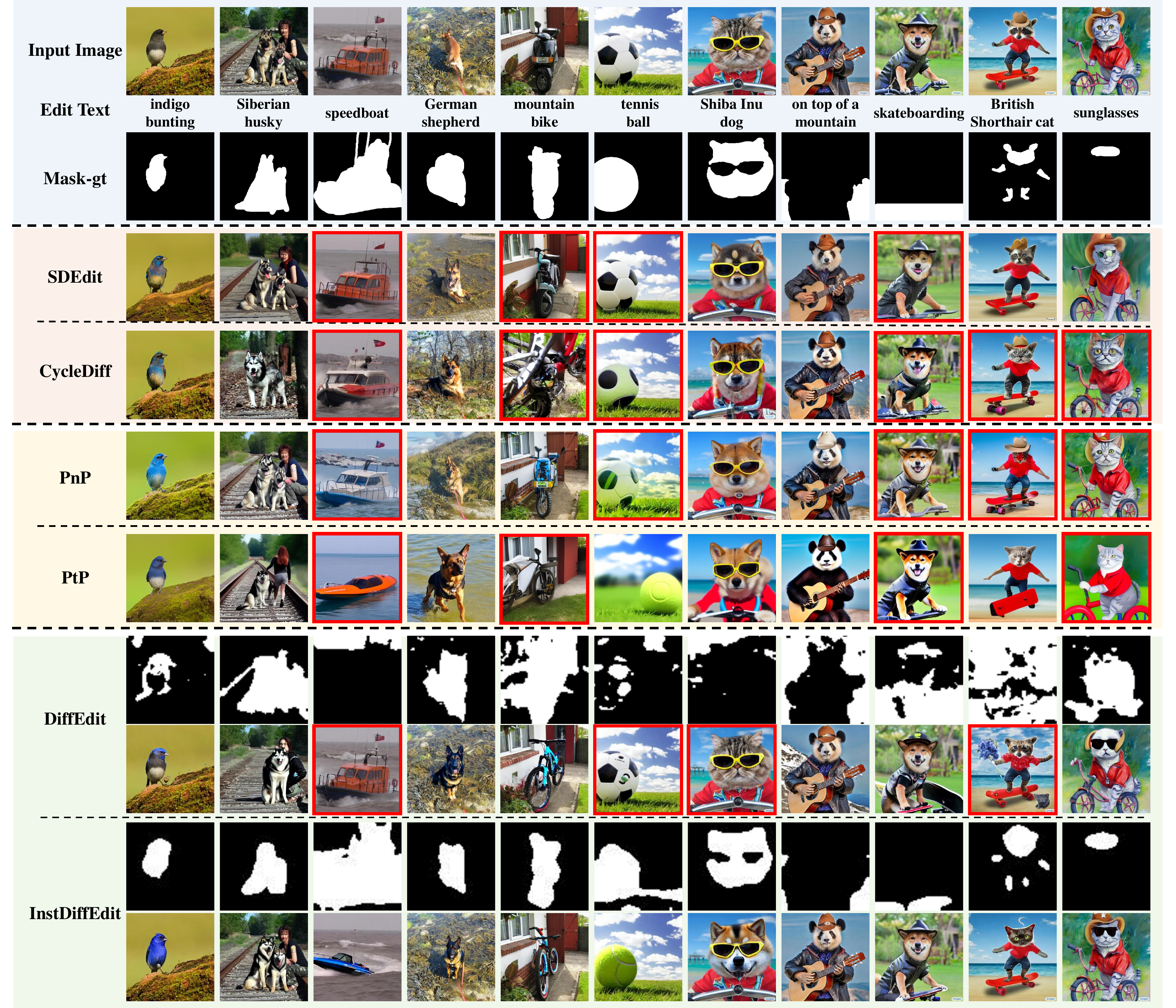} 
\caption{Visualizations of the generated masks and edited images of InstDiffEdit and the compared methods. Compared with DiffEdit, the masks of InstDiffEdit are closer to the human-labeled ones. Moreover, the comparisons with the latent-based and attention-based approaches also show the merit of the instant mask in our InstDiffEdit. The red boxes refers to failed editions.}
\label{fig.visions}
\end{figure*}

\subsubsection{\textbf{Qualitative Analysis}}
To obtain deep insight into InstDiffEdit, 
we visualize the editing results of our InstDiffEdit and other compared methods on Editing-Mask, as shown in Fig. \ref{fig.visions}. 
It can be first seen that both latent-based and attention-based approaches lack explicit constraints on the area to edit, which may result in unexpected generations. For instance, in the case of the \emph{``German Shepherd''} image in the 4th column, DiffEdit and InstDiffEdit successfully modify the object while preserving the background, while other mask-free methods obviously change the background. However, a noteworthy disparity exists between the generated masks of DiffEdit and the human-labeled masks. Specifically, the masks produced by DiffEdit are somewhat inaccurate, and exhibits peculiar shape outlines. In contrast, our generated masks are significantly superior to those generated by DiffEdit, leading better editing results. For instance, in the case of \emph{``speedboat''} image in the 3rd column, our mask accurately encompasses the primary object \emph{``boat''}, whereas the mask generated by DiffEdit is non-representative. Consequently, our approach achieves successful editing, whereas DiffEdit fails to do so. These results are consistent with IOU performance presented in Tab. \ref{tab:main-results}.


\section{Conclusion}
In this paper, we propose a novel and efficient method, called InstDiffEdit for diffusion-based semantic image editing. As an plug-and-play component, InstDiffEdit can be directly applied to most diffusion models without any additional training or human intervention. Experimental results not only demonstrate the superior performance of InstDiffEdit in semantic image editing tasks, but also confirm its superiority in computation efficiency, \emph{e.g.}, up to 5 to 6 times faster than DiffEdit. 
\section{Acknowledgments}
This work was supported by National Key R\&D Program of China (No.2023YFB4502804) , the National Science Fund for Distinguished Young Scholars (No.62025603), the National Natural Science Foundation of China (No. U22B2051, No. U21B2037,  No. 62176222, No. 62176223, No. 62176226, No. 62072386, No. 62072387, No. 62072389, No. 62002305 and No. 62272401),  the Key Research and Development Program of Zhejiang Province (No. 2022C01011), the Natural Science Foundation of Fujian Province of China (No.2021J01002,  No.2022J06001), and partially sponsored by CCF-NetEase ThunderFire Innovation Research Funding (NO. CCF-Netease 202301).


\bibliography{aaai24}

\clearpage

\setcounter{figure}{0}
\renewcommand\thefigure{\Roman{figure}}  
\renewcommand\thetable{\Roman{table}}  
\renewcommand\theequation{\Roman{equation}} 
\setcounter{figure}{0}  
\setcounter{table}{0}  
\setcounter{equation}{0}

\appendix
\section*{APPENDIX}
\subsection{Evaltion Dataset}

\textbf{InstDiffEdit} mainly conducts experiments on 3 datasets, namely \emph{ImageNet}, \emph{Imagen} and \emph{Editing-Mask}. In this section, we will introduce these datasets in details.

\begin{figure} [tp]
\centering
\includegraphics[width=1\linewidth]{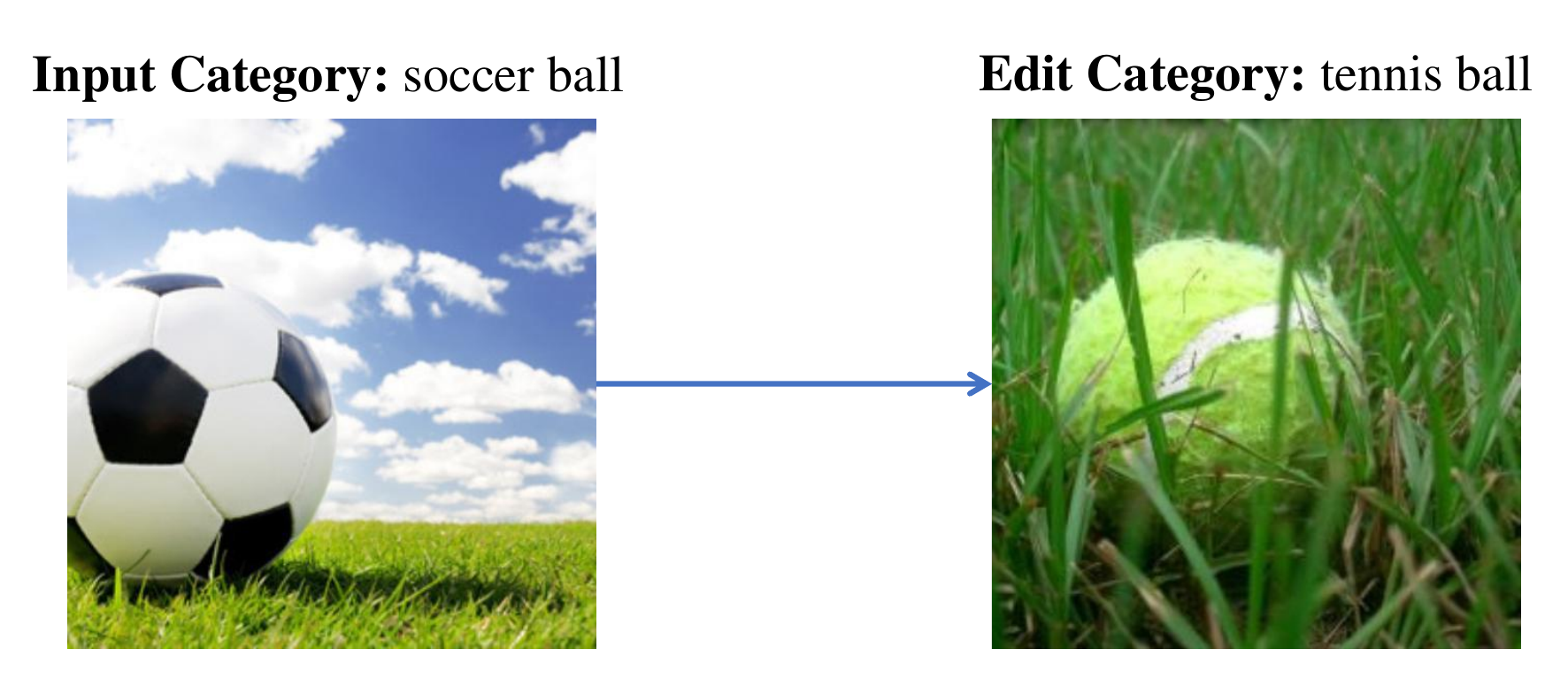} 
\caption{ A instance in ImageNet dataset. The category "tennis ball" is used to edit a image of category "soccer ball".}
\label{fig:ImageNetdataset}
\end{figure}

\begin{figure} [tp]
\centering
\includegraphics[width=1\linewidth]{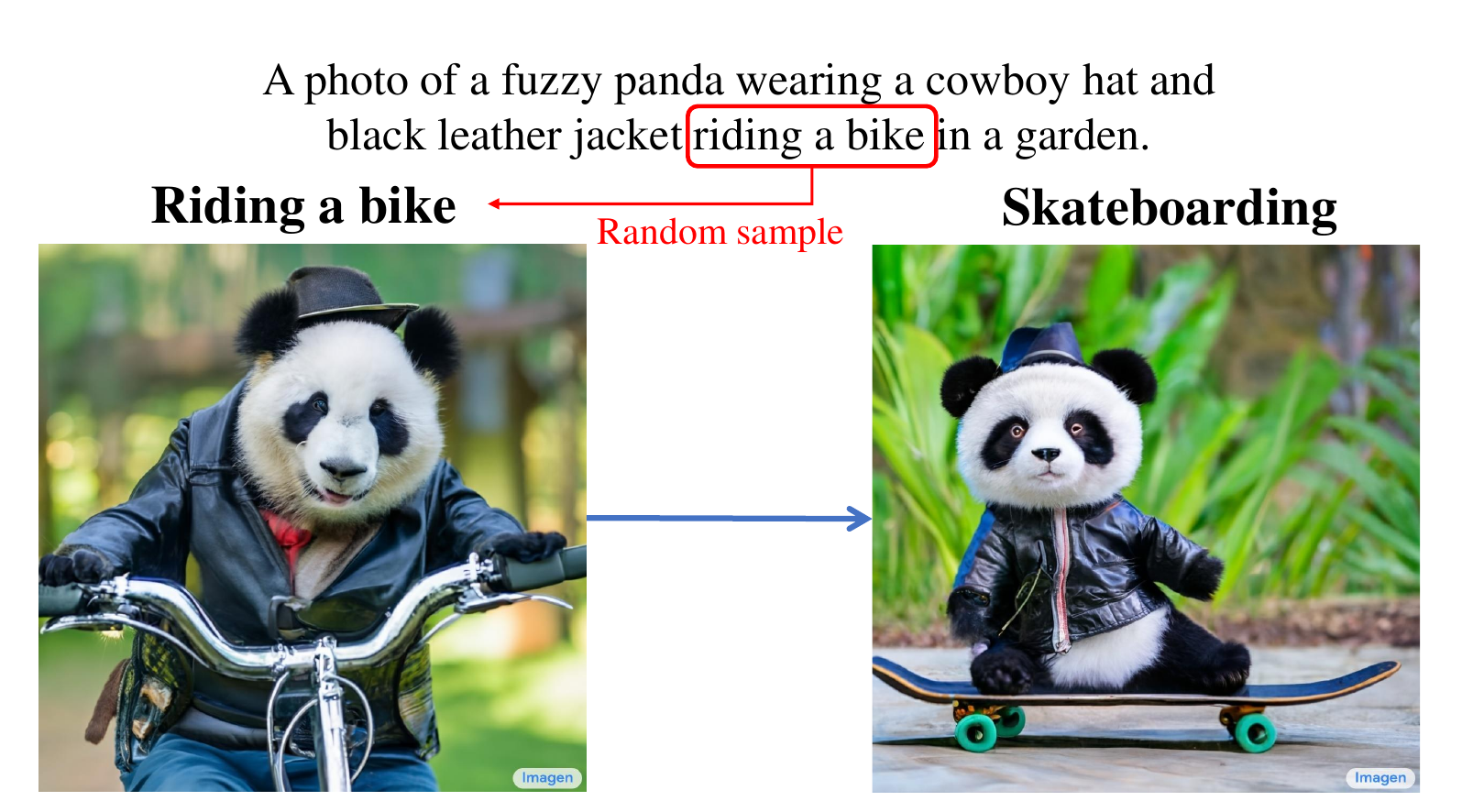} 
\caption{ A instance in Imagen dataset. We construct an editing text by replacing parts of the original sentence with short phrases, such as "skateboarding"
.} 
\label{fig:Imagendataset}

\end{figure}

\begin{figure*} [t]
\centering
\includegraphics[width=1\linewidth]{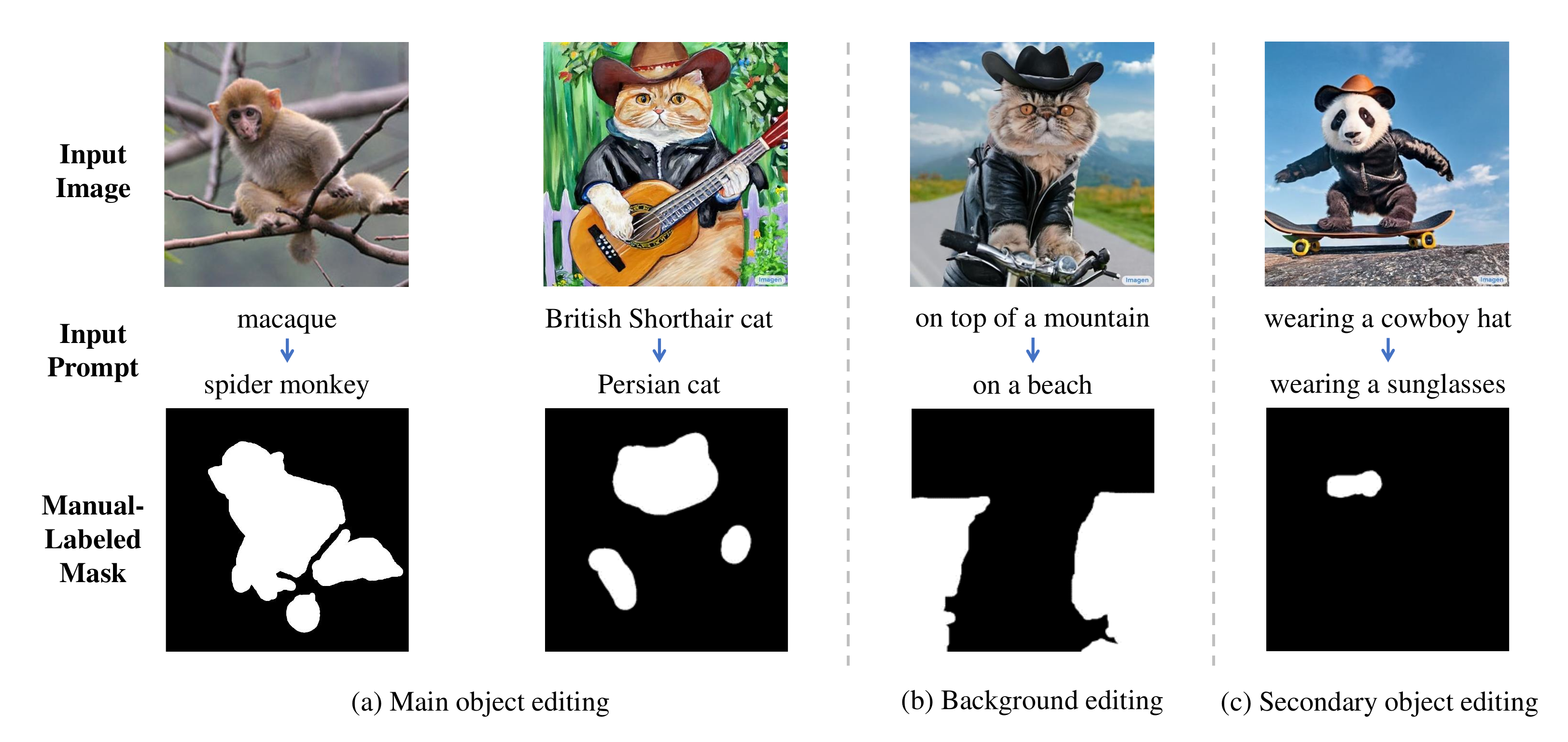} 
\centering
\caption{Different editing instance on Editing-Mask dataset, including main object, background 
 and secondary object editing.}
\label{fig:editing}
\end{figure*}

\begin{itemize}
 \item \textbf{ImageNet} We use the evaluation settings introduced in  \emph{FlexIT} \cite{couairon2022flexit}, and the dataset includes 1092 images, covering 273 categories. Each sample in the dataset comprises an image and a corresponding text category label. In this dataset, editing is performed by using a similar category label as the editing text to complete the task. For example, as shown in Fig. \ref{fig:ImageNetdataset}, the category label "tennis ball" is utilized to edit an image categorized as "soccer ball".

 \item \textbf{Imagen}  
To evaluate semantic editing performance, we construct a dataset using the generations from \emph{Imagen}, which consists of 360 images with corresponding generation prompts. As shown in Fig. \ref{fig:Imagendataset}, a instance includes a text "A photo of a fuzzy panda wearing a cowboy hat and black leather jacket riding a bike in a garden." and a corresponding image. For local editing tasks, we randomly selected a short text that was not present in the original generation prompt as the editing text. For example, we replaced "riding a bike" with "skateboarding" to obtain a editing sample.

 \item \textbf{Editing-Mask} 
We propose a new dataset termed \emph{Editing-Mask}, which features manual annotations of regions in an image that require editing to evaluate the performance of local editing task. We randomly selected 200 images from the Imagen and ImageNet datasets, with text of varying lengths included in each instance. The mask for each image is labeled by human with the minimum area required for successful editing based on the edit text provided. As illustrated in Fig. \ref{fig:editing}, the dataset covers editing operations in three directions: main object editing, secondary object editing and background editing. For example, changing "macaque" to "spider monkey" requires editing the main object while preserving the background area in the image. Overall, each instance in the dataset includes an input image, original and edit text, and a manually-labeled mask reference for the edit text.
\end{itemize}

\subsection{Evaltion Metrics}

In local editing task, the generated image is required to conforms to the semantics of the edited text while preserving the original image information, such as the background. 

Our evaluation of editing methods in local editing tasks involves two aspects. Firstly, we use \textbf{LPIPS} to measure the similarity between the generated images and the original images in the Imagen and ImageNet datasets. A smaller LPIPS value indicates that the edited image retains more information from the original image, which aligns with the goal of editing tasks to preserve as much information as possible.
Secondly, we evaluate the performance of editing methods in local editing tasks using a category FID called \textbf{CSFID}. This metric calculates the FID between the generated images and the original images that belong to the same category, which reflects whether the edited image matches the editing category. However, since there is no corresponding category in the Imagen dataset, we use FID to measure the image quality and \textbf{CLIPScore} to assess the effectiveness of the editing. CLIPScore measures the similarity between the edited images and the corresponding edited text, indicating whether the edit has taken effect.
 
The aforementioned metrics are primarily introduced to implicitly measure the quality of the generated images, rather than the performance of the editing methods. To explicitly reflect the local editing ability of the models, we propose a new dataset that includes several metrics. Firstly, for the mask-based method in our paper, we measure the mask accuracy using the \emph{Intersection over Union} (\textbf{IoU}) between the generated mask $M_{gen}\in \{0,1\}$ and the manually labeled mask $M_{gt}\in \{0,1\}$, which can be expressed as:

\begin{equation}
    IoU=\frac{M_{gen}\cap M_{gt}}{M_{gen}\cup M_{gt}}
\end{equation}

Furthermore, we propose two additional metrics, namely $C_m$ and $C_{non}$, to measure the local editing performance of all the methods. These metrics indicate the rate of change of pixels in the mask and non-mask regions, respectively, based on the manually labeled masks. They are represented by:
\begin{equation}
    C_{m}=\frac{{\sum}_{(i,j)\in M_{gt}} p(i,j) }{255 \cdot {\sum}_{(i,j)} M_{gt}(i,j)}
\end{equation}
\begin{equation}
    C_{non}=\frac{{\sum}_{(i,j)\notin M_{gt}} p(i,j) }{255 \cdot {\sum}_{(i,j)} (1-M_{gt}(i,j))}
\end{equation}
where, $p(i,j)\in [0,255]$ represents the pixel value changed in point $(i,j)$ between input image and generation image.
Finally, we use the ratio of $C_m$ to $C_{non}$ to measure the relative value of the local editing ability of different methods.

\subsection{Implementation Details of Baseline models} 

\textbf{SDEdit} We implement the latent-based editing method with the framework of Stable Diffusion.  By leveraging the stability of the diffusion process, SDEdit regulates the extent of preservation of the original image information via the introduction of noise intensity $r$. Textual information is subsequently incorporated into the generation process via classifier-free guidance (CFG). We set the CFG scale $\lambda$ to 7.5 by default and measure the performance of SDEdit under different noise strengths $r$.

\textbf{CycleDiffusion} We implement it with the official project with  default parameters.(\url{https://github.com/ChenWu98/cycle-diffusion}). By proposing a superior inversion method, CycleDiffusion can obtain a more accurate latent representations of the images, which is the basis for latent-based and attention-based image editing method. 
Considering the importance of the text used by the model for inversion, and based on the fact that using short text as a prompt for generation is very inefficient, we have made adaptive improvements to the method for some datasets. As shown in Fig. \ref{fig:Imagendataset}, we obtain the prompt of the Imagen dataset by randomly selecting a phrase in the long text and then replacing it. After improvement, we replace the  phrase directly in place of the original phrase in the long text. In this way, we can get the long prompt containing the replacement phrase, and use it as the edit text of CycleDiffusion. 
In the editing-mask dataset, part of the data from Imagen is consistent with the above settings. As for the data from ImageNet, we first obtain the image caption through BLIP \cite{li2022blip}, and then manually add the category name which is the prompt in the original setting to the caption, for getting a long prompt that contains more information. Since the improvement on ImageNet involves too much manual participation and is time-consuming, we retain the settings of ImageNet, as shown in Fig. \ref{fig:ImageNetdataset}.



\textbf{PnP} We implement PnP with the official project (\url{https://github.com/MichalGeyer/PnP-diffusers}) with default parameters. 
By injecting the attention information retained in the image reconstruction process into the generation process, PnP can ensure the structural consistency between the generated images and the input images.

\textbf{PtP} We first use the inversion formula of DDIM\cite{song2020denoising} with 50 inference steps to obtain the latent representation of the real image, and then generate it through official project (\url{https://github.com/google/prompt-to-prompt}) and parameters. 
Similar to PnP, PtP is also an attention-based method, which requires the use of inversion operations. At the same time, considering that the PtP method requires the same text length before and after editing, we further adapt the dataset settings used by CycleDiffusion. On the Imagen and Editing-Mask datasets, we replicate the last word (usually a noun) of a shorter phrase until it matches the length of the longer phrase. For the ImageNet dataset, due to the large amount of data and the poor performance of PtP under the setting of using the category name as the prompt, we did not conduct this experiment.

\textbf{DiffEdit} is a new method for text-driven image editing based on Stable Diffusion models with classifier-free guidance(CFG), which calculates semantic differences and generates masks for automatic editing. We performed a lightweight hyperparameter search to optimize the best trade-offs in the matrix on three datasets. Since the official DiffEdit code is not publicly available, we implement it referred to \url{https://github.com/johnrobinsn/diffusion_experiments/}.  Additionally, to ensure fair comparison, we also used the masks obtained from DiffEdit as input for the inpainting method to generate images. Note that results are based on a non-official re-implementation and do not fully represent the performance of the original paper.

\subsection{Supplementary experiments}

We report the results of single and multiple objects settings on Tab. \ref{tab.mutial}. It can be also seen that InstDiffEdit has the ability to handle multiple objects in images, and it is consistently better than DiffEdit under all settings.
\begin{table}[t]
    \centering
    \setlength\tabcolsep{4pt}
    \renewcommand\arraystretch{0.8}
    \begin{tabular}{c|c|cccc}
    \toprule
    {\textbf{Mode}} & {\textbf{Method}}  & {{$C_m$}$\uparrow$ } & {{$C_{non}$}$\downarrow$ }  & {{$C_{m}$/$C_{non}$}$\uparrow$} & {\textbf{IoU}$\uparrow$}\\
    \toprule

    \multirow{2}*{\textbf{Single}} 
      & DiffEdit & 0.171 & 0.069 & 2.489 & 36.1\\
    ~ & InstDiffEdit & \textbf{0.235} & \textbf{0.043} & \textbf{5.478} & \textbf{57.1} \\
    \midrule
    \multirow{2}*{\textbf{Multi}} 
      & DiffEdit & 0.158 & 0.085 & 1.865 & 34.5 \\
    ~ & InstDiffEdit & \textbf{0.204} & \textbf{0.048} & \textbf{4.252} & \textbf{49.5}\\
    \midrule
    \multirow{2}*{\textbf{ALL}} 
      & DiffEdit & 0.168 & 0.072 & 2.314 & 35.7\\
    ~ & InstDiffEdit & \textbf{0.227} & \textbf{0.044} & \textbf{5.157} & \textbf{55.3} \\
    \bottomrule
    \end{tabular}
    \caption{The result of Single object and Multiple objects.}
    \label{tab.mutial}
\end{table}

\subsection{More visualization results}
\label{B}
As shown in Fig. ~\ref{fig.visions} and Fig. ~\ref{fig.visions2},  we  illstruate more editing results of all the methods from ImageNet and Imagen datasets, which indicating InstDiffEdit can always achieve the best editing performance.
\begin{figure*} [btp]
\centering
\includegraphics[width=1\linewidth]{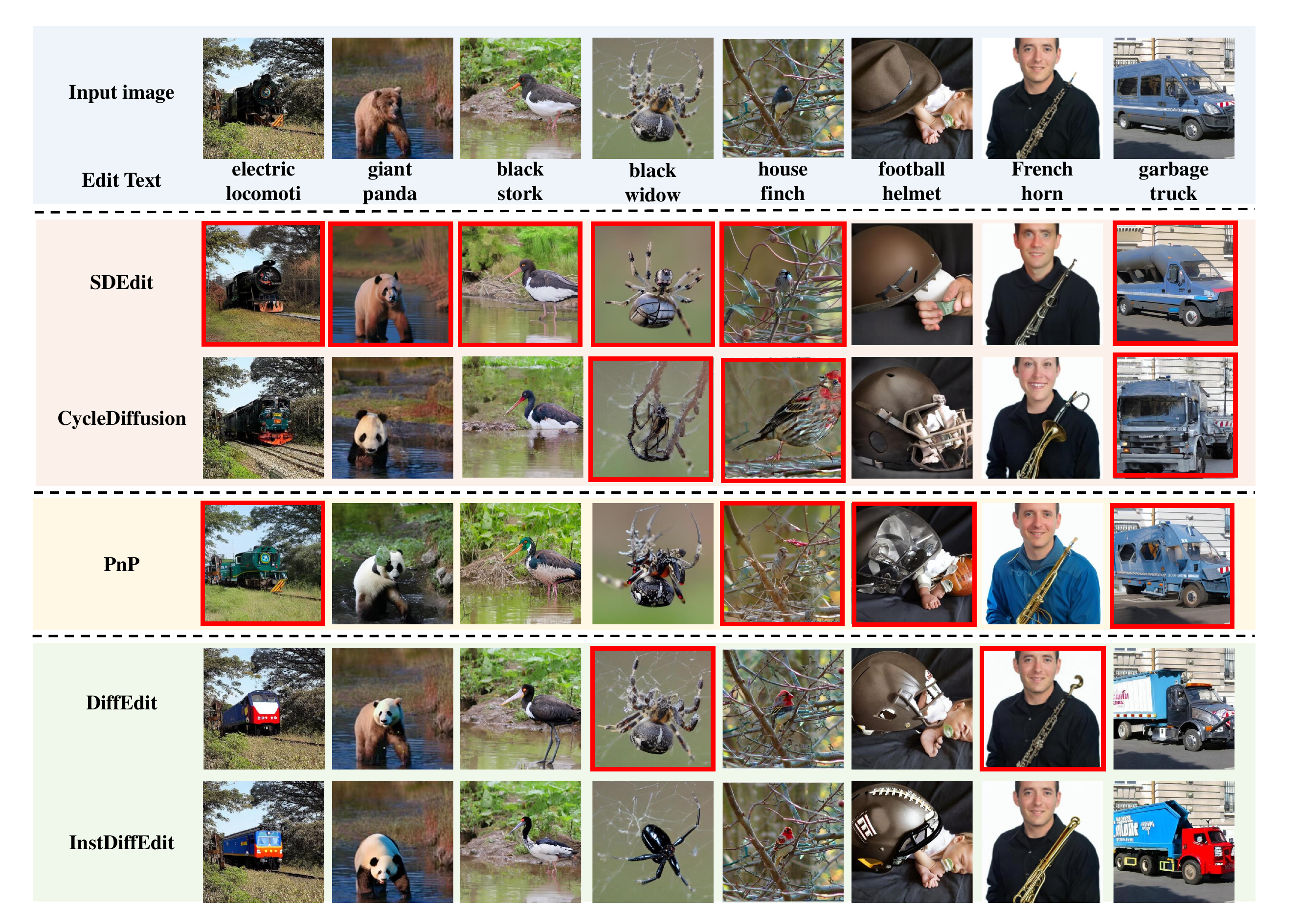} 
\caption{More Visualizations of the edited images of all the methods in ImageNet dataset. InstDiffEdit enables successful editing with minimally modified regions on a wider range of image categories. The red boxes refers to failed editions.}

\label{fig.visions}
\end{figure*} 

\begin{figure*} [btp]
\centering
\includegraphics[width=1\linewidth]{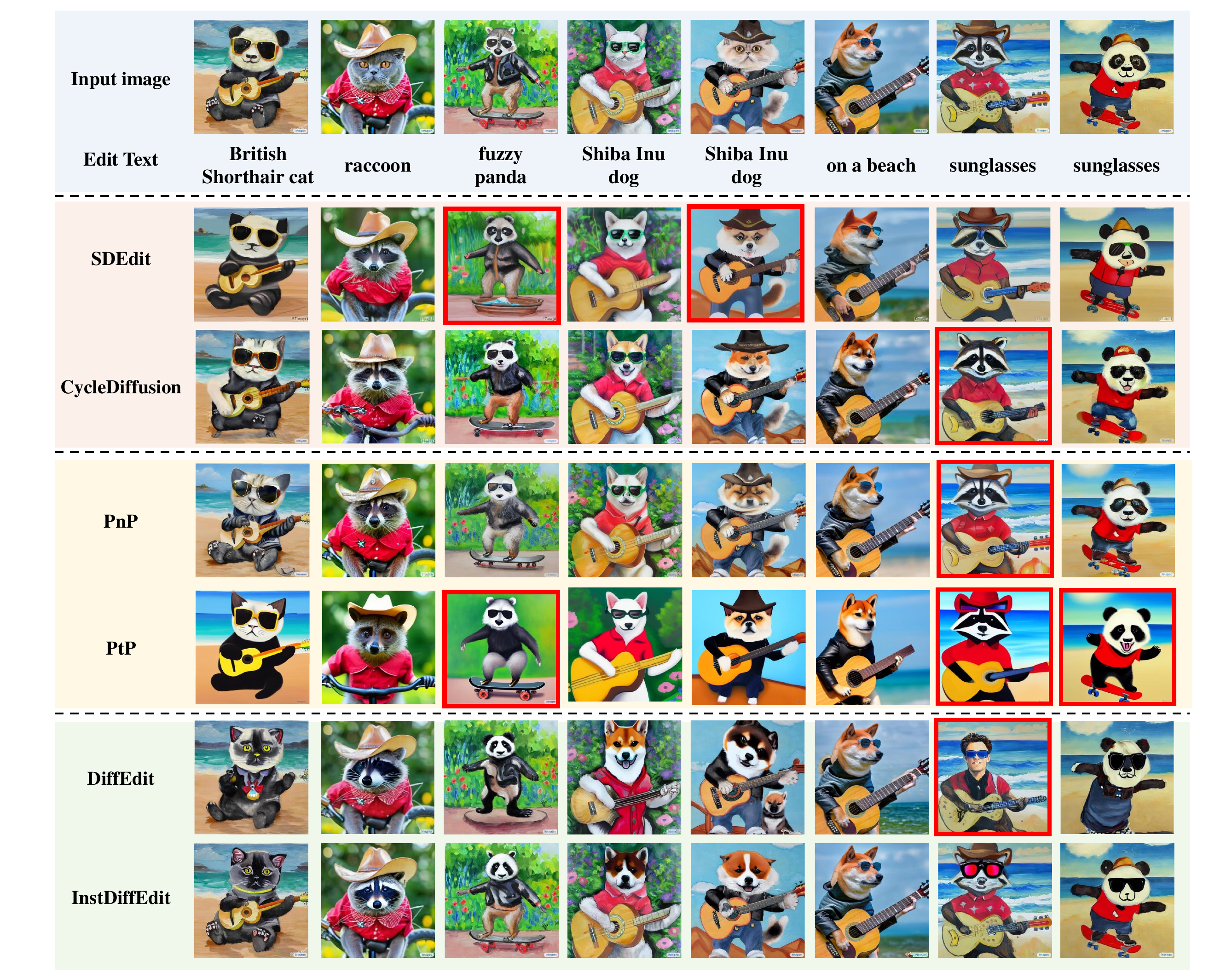} 
\caption{More Visualizations of the edited images of all the methods in Imagen dataset. InstDiffEdit enables successful editing with minimally modified regions on a wider range of image categories. The red boxes refers to failed editions.}
\label{fig.visions2}
\end{figure*} 

\end{document}